\documentclass{article}
\usepackage{ijcai19}

\usepackage{times}
\usepackage{soul}
\usepackage{url}
\usepackage[hidelinks]{hyperref}
\usepackage[utf8]{inputenc}
\usepackage[small]{caption}
\usepackage{graphicx}
\usepackage{amsmath}
\usepackage{booktabs}
\usepackage{algorithm}
\usepackage{algorithmic}
\title{CompactNet: Platform-Aware Automatic Optimization for Convolutional Neural Networks}

\author{
Weicheng Li$^1$\and
Rui Wang$^1$\footnote{Contact Author}\and
Zhongzhi Luan$^1$\and
Di Huang$^1$\and
Zidong Du$^2$\and
Yunji Chen$^2$\And
Depei Qian$^1$
\affiliations
$^1$School of Computer Science and Engineering, Beihang University, Beijing, China\\
$^2$Institute of Computing Technology, Chinese Academy of Sciences, Beijing, China\\
\emails
\{liweicheng, wangrui, luan.zhongzhi, dhuang, depeiq\}@buaa.edu.cn,
\{cyj, duzidong\}@ict.ac.cn
}

\begin{document}

\maketitle

\begin{abstract}
Convolutional Neural Network (CNN) based Deep Learning (DL) has achieved great progress in many real-life applications. Meanwhile, due to the complex model structures against strict latency and memory restriction, the implementation of CNN models on the resource-limited platforms is becoming more challenging.  This work proposes a solution, called CompactNet\footnote{Project URL: \url{https://github.com/CompactNet/CompactNet}}, which automatically optimizes a pre-trained CNN model on a specific resource-limited platform given a specific target of inference speedup. Guided by a simulator of the target platform, CompactNet progressively trims a pre-trained network by removing certain redundant filters until the target speedup is reached and generates an optimal platform-specific model while maintaining the accuracy. We evaluate our work on two platforms of a mobile ARM CPU and a machine learning accelerator NPU (Cambricon-1A ISA) on a Huawei Mate10 smartphone. For the state-of-the-art slim CNN model made for the embedded platform, MobileNetV2, CompactNet achieves up to a 1.8x kernel computation speedup with equal or even higher accuracy for image classification tasks on the Cifar-10 dataset.
\end{abstract}

\section{Introduction}

CNN is a dominant representative in DL, delivering outstanding performance on many computer vision tasks such as image classification and object detection. However, CNN-based DL applications are typically too computationally intensive to be deployed on resource-limited platforms such as smartphones where latency is very sensitive to the user experience.

At present, almost all CNN structures are designed without considering backend platforms. It is hard for a single network to run across all platforms in an optimal way due to the different hardware architecture characteristics. For example, the fastest CNN model on a desktop GPU may not be the fastest one on an embedded domain-specific accelerator with the same accuracy. However, manually crafting CNN structures for a given target platform requires deep knowledge about details of the backend platform, including the toolchains, configuration, and hardware architecture, which are usually unavailable.

Moreover, a large number of works have focused on DL model optimization such as weights-sparsity and pruning techniques. In spite of promising optimization results, they are not platform-aware and cannot generate optimal models across all platforms from servers or desktops to resource-limited devices.

In this work, we propose a platform-aware solution, called CompactNet, to address the aforementioned issues. It automatically generates optimal platform-specific CNN models with a certain speedup target guaranteed. CompactNet (see Figure 1) is driven by a platform simulator that collects data from real backend hardware and simulates the latency of a model trimmed by removing certain redundant filters in certain convolutional layers. Guided by the simulated latency, the searching loop can generate the best-trimmed model satisfying the target speedup with the highest accuracy. Since the latency can be simulated on any platform that supports common DL algorithms, this solution supports any platform without detailed knowledge of the platform itself.

\begin{figure}
\centering
\centerline{\includegraphics[width=\columnwidth]{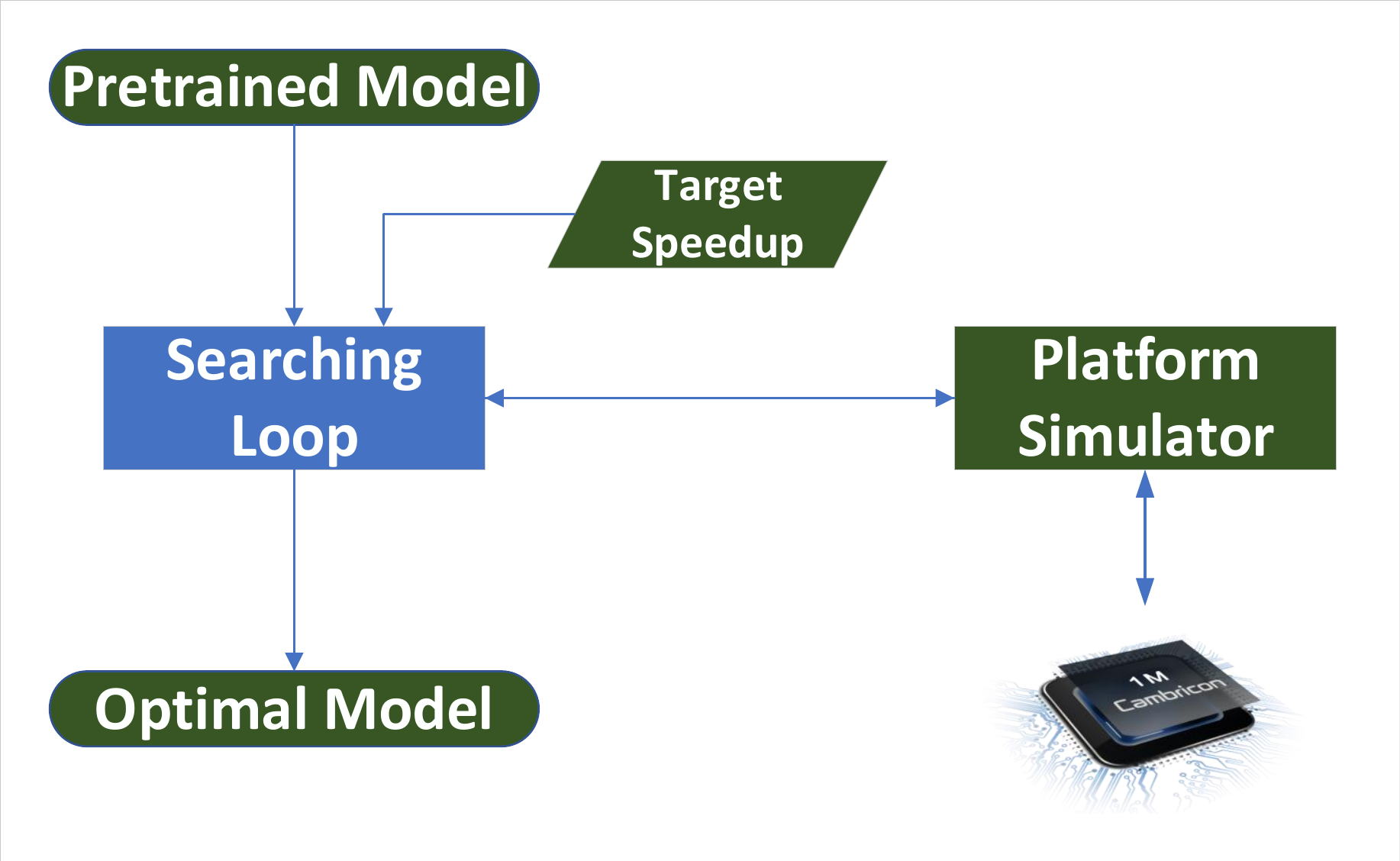}}
\caption{Framework of CompactNet.}
\end{figure}

The main contributions of this paper are:
\begin{itemize}
\item \textbf{A general and adaptable platform simulator that guides the whole approach}. This simulator can collect data from any platforms that support common DL algorithms and precisely simulate the latency of them.
\item \textbf{An automatic optimizing approach with a certain speedup target guaranteed.} We trim the entire filters (output channels) instead of individual weights of each layer until the speedup target is satisfied. This makes our approach more direct and purposive. 
\item \textbf{A platform-aware optimizing approach that generates platform-specific optimal CNN models without knowing the backend hardware architecture details.} The experiments show that optimal models for different platforms with the same speedup generated by our approach have different structures. An optimal model for a certain platform cannot achieve the same speedup on another as the optimal model specifically for that different platform. The intrinsic reason for this should be deep in the hardware architecture characteristics and our work can be considered to provide a black-box to interact with them.
\end{itemize}

\section{Related Work}

In recent years, a large number of works~\cite{1} aiming to optimize CNN models have achieved great success. Most of the works can be divided into two main categories.

First, lots of famous works adopt pruning techniques~\cite{2}~\cite{4}. These approaches focus on removing the redundant weights to sparsify the filters in the model. And they can be further divided into weight-level~\cite{5}~\cite{6}, vector-level~\cite{7}, kernel-level~\cite{8} and group-level~\cite{9}~\cite{10}. Unfortunately, not all the platform can fully take advantage of such sparse data structure~\cite{11} and therefore, there is no guarantee on reducing the latency. Other works~\cite{12}~\cite{13}, in contrast, consider removing entire filters, which shows a more conspicuous speedup. The main issue of these approaches is that they are not automatic or platform-aware. It means that the number of removed filters needs to be set manually since different backend platforms may have different optimal options. ADC~\cite{15} proposes using reinforcement learning and MorphNet~\cite{29} leverages the sparsifying regularizers to decide the compression rates. AdaptNet~\cite{16} uses direct metrics as guides for adapting DL models to mobile devices given a specific resource budget. Our CompactNet addresses the same issue in a different way by removing certain redundant filters according to the simulated latency data of the backend platforms in order to satisfy the target speedup.

Another way to optimize CNN models is focusing on network structure. MobileNets~\cite{17}~\cite{18} SqueezeNet~\cite{19} and~\cite{20} are typical examples of this kind. They are all general designs to build more efficient CNN models by removing the FC layer, using multiple group convolution or proposing depth-wise convolution. There’s no doubt that such works have achieved great success in saving resources and reducing latency. However, they are not designed for specific platforms and our experiments show that deployed on different backend platforms, they still have a significant optimizing space for kernel computation speedup via our CompactNet.

Besides, other approaches based on low-rank approximation~\cite{21}~\cite{22} use matrix decomposition to reduce the number of operations. The motivation behind such decomposition is to find an approximate matrix that substitutes the original weights.  And others like~\cite{23}~\cite{24} focus on the data type and significantly reduce the latency by quantization. All those works are stand-alone optimization and can be considered as complements to our CompactNet. 

\section{CompactNet}

Different from the solutions in the literature, we propose a platform-aware optimization, called CompactNet, that automatically trims a pre-trained CNN model given a certain target speedup while maintaining the accuracy. CompactNet is driven by a platform simulator based on the actual target platform. Guided by the simulated latency data, CompactNet generates platform-specific optimal models by progressively removing redundant filters without the requirement of expertise of the platform itself and guarantees the target speedup.

\subsection{General and Adaptable Platform Simulator}

The platform simulator is the foundation of our work. As guided by the latency data simulated on real hardware platforms by the simulator, our CompactNet can be so-called platform-aware and is able to trim a CNN model purposively given a certain speedup target.

Different from traditional hardware simulators, the proposed one here does not focus on the architecture or mechanism of the platform. Instead, it directly simulates the kernel computation latency of a CNN model on the real backend platform. The implementation of the simulator is not complicated so long as the platform supports common DL algorithms. Taking MobilenetV2~\cite{18} (see Table 1) as an example, we implement the 18 layers respectively including the first Conv2D layer and each of the following bottleneck layers on the target backend platform. Then we continuously change the number of input channels and filters (output channels) of each layer. Thus we can collect the latency data of each layer with any number of input and output channels. A simple example that the data collected by the simulator is shown in Figure 2.

\begin{figure*}
\centering
\centerline{\includegraphics[width=18.5cm]{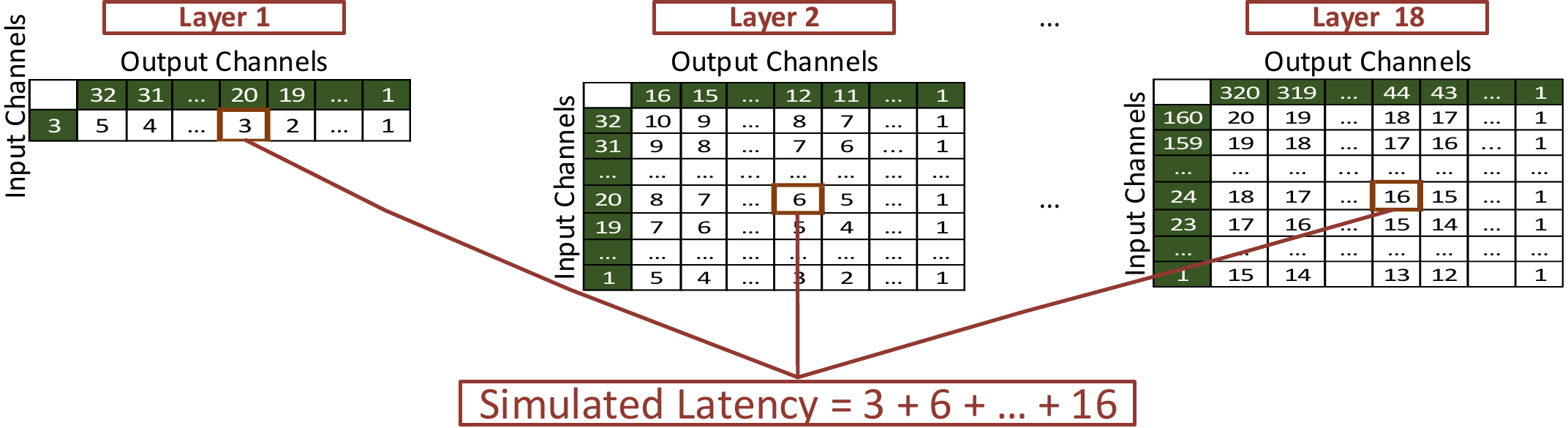}}
\caption{Illustration of collecting latency data by the simulator and using it to simulate the kernel computation of a certain trimmed model as the one in Table 2.}
\end{figure*}

\begin{table}
\centering
\begin{tabular}{lrrr}  
\toprule
Layers  & Input Channels & Output Channels & $n$ \\
\midrule
Conv2D       & 3  & 32 & 1      \\
Bottleneck       & 32  & 16 & 1      \\
            & 16  & 24 & 2       \\
            & 24  & 32 & 3       \\
            & 32  & 64 & 4       \\
            & 64  & 96 & 3       \\
            & 96  & 160 & 3       \\
            & 160 & 320 & 1       \\
Conv2D       & 320  & 1280 & 1       \\
            & 1280  & k & 1       \\
\bottomrule
\end{tabular}
\caption{Original MobileNetV2 architecture where $n$ means the repeat times of the layer.}
\end{table}

\begin{table}
\centering
\begin{tabular}{lrrr}  
\toprule
Layers  & Input Channels & Output Channels & $n$ \\
\midrule
Conv2D       & 3  & 20 & 1      \\
Bottleneck       & 20  & 12 & 1      \\
            & 12  & 20 & 2       \\
            & 20  & 24 & 3       \\
            & 24  & 12 & 4       \\
            & 12  & 22 & 3       \\
            & 22  & 24 & 3       \\
            & 24 & 44 & 1       \\
Conv2D       & 44  & 1280 & 1       \\
            & 1280  & k & 1       \\
\bottomrule
\end{tabular}
\caption{Example of a trimmed MobileNetV2 model architecture by removing certain redundant filters across layers.}
\end{table}

With the collected data, we can simulate the latency of kernel computation of a certain trimmed model (after removing some filters across layers like in Table 2) by simply summing up the latency of each layer with the trimmed number of input and output channels (see Figure 2). Such simulated latency is well approximates the real latency of the model according to experiments (see Section 4.3). Then in the searching loop, we can decide whether the trimmed model satisfy the target speedup or not. The details of the searching loop will be introduced in the next part.

Our simulator is general and adaptable. It can collect latency data of any kind of CNN model on any platform by simply implementing each layer of the model on the target platform and executing with any number of input and output channels.

\subsection{Automatic and Platform-Aware Searching Algorithm}

In this part, we introduce the searching loop that generates the optimal trimmed model given a target speedup and its relationship with the platform simulator. The whole process is shown in Figure 3-left.

\begin{figure*}
\centering
\centerline{\includegraphics[width=18.5cm]{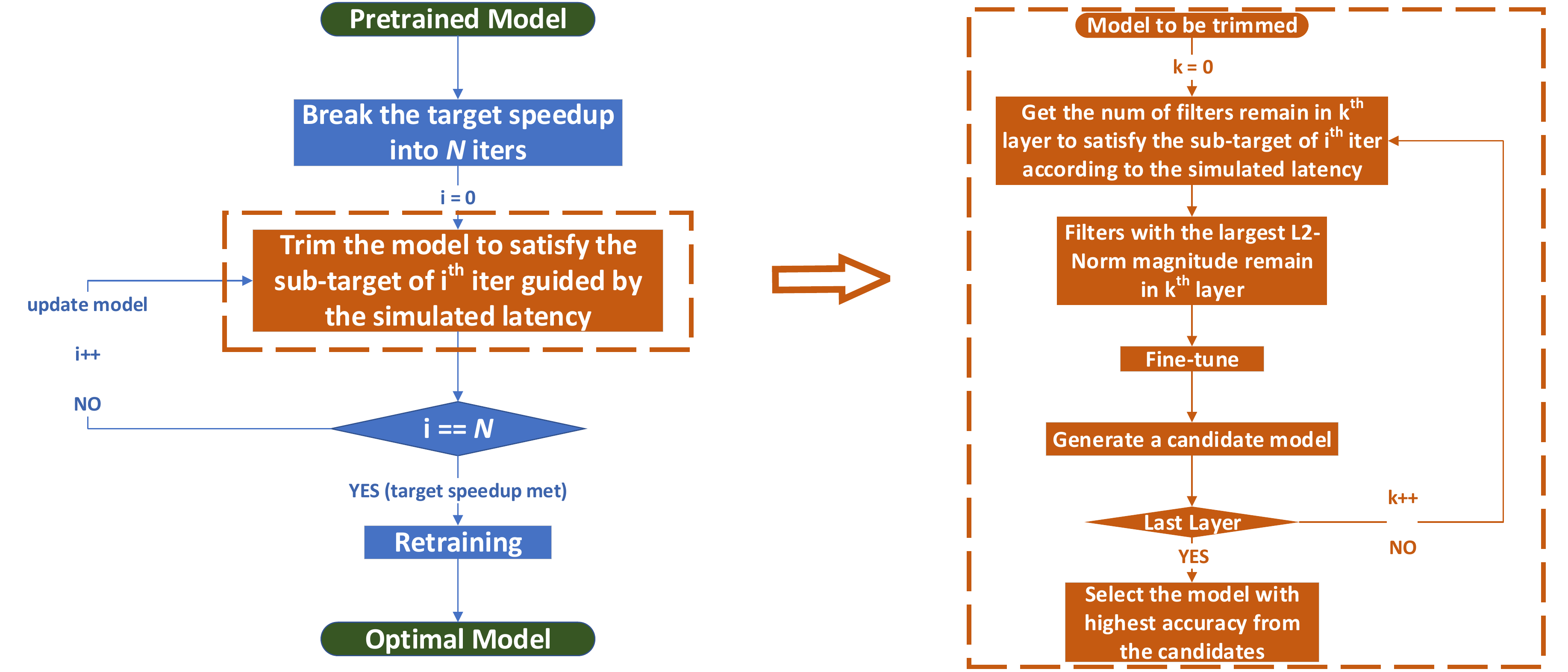}}
\caption{Searching Loop inside CompactNet. We break the target speedup into $N$ iterations and each of the iterations has a sub-target. Overall (left), we progressively trim the model until finishing the last iteration when the target speedup is reached. For each iteration (right), guided by the simulated latency, we generate $k$ candidate models satisfying the sub-target where $k$ is the number of trimmable layers in the model and select the one with the highest accuracy. In one iteration, we essentially trim only the one layer that loses the least accuracy to satisfy the sub-target.}
\end{figure*}

To trim the pre-trained model and satisfy the target speedup progressively, we first break the optimization into $N$ iterations, where $N$ is one of the key hyperparameters in this work. So in each iteration, we have a sub-target of speedup. Here we use another important hyperparameter – the decay of sub-target of speed up $d$ which is similar to the decay of learning rate in DL model training. It enables us to trim the model more carefully during the searching to avoid removing imperative filters. Therefore, the sub-target of speedup for each iteration $T_{subi}$ is computed by Eq. 1 where $T_{init}$ is the initial target of speedup without considering the decay. Thus the final target speedup $T_{final}$ can be computed by Eq. 2. In fact, while doing real tasks, we need to keep $T_{final}$ in mind and inversely compute $T_{init}$ first. For each iteration, guided by the simulated latency, we trim the model to satisfy $T_{subi}$ and the trimmed model is used in the next iteration. When we finish the last iteration and all the sub-targets are satisfied, we get a new model that meets the final speedup target. Then we retrain it and generate the final optimal model.

\begin{equation}
T_{subi}=\ \frac{T_{init}}{N}\ast\ d^i,\ i=0,\ldots,N-1
\end{equation}

\begin{equation}
T_{final}=\ \sum_{i=0}^{N-1}T_{subi}
\end{equation}

With the platform simulator, the whole searching loop does not need to be deployed on or interact with the target platform itself. All the latency data required could be simulated before. Each iteration in the loop takes the simulated latency data and trims the model to satisfy $T_{subi}$. Since different platforms have different latency data, the final optimal model is platform-specific. Obviously, the trimming step is important in each iteration. In the next part, we will discuss the details of it.

\subsection{Trimming Approach}

In this part, we introduce the core step in the searching loop to trim the model progressively according to the simulated latency data. The inside process of this trimming step is shown in Figure 3-right.

During each iteration in the searching loop, we generate $k$ candidate models where $k$ is the number of trimmable layers in the CNN model and select the one with the highest accuracy as an update for the next iteration. For each iteration, to generate the candidate models, the trimming approach goes through each of the $k$ trimmable layers in the model. For each layer, guided by the simulated latency data, we reduce the number of filters (output channels) one by one until the trimmed model latency satisfies the sub-target speedup ($T_{subi}$ in Eq.1) of this iteration. Note that when the number of filters in one layer is modified, the number of input channels of the next layer should also be modified accordingly. 

After we know how many filters should remain in current layer to satisfy the sub-target. We need to decide which filters should be preserved. There are several studies like~\cite{25} discussing the influence of certain filters in a model and can be used to choose filters here. We simply choose the magnitude-based method that filters with the largest L2-norm magnitude remain.

Then we fine-tune the model with the trimmed layer to restore the accuracy and generate the fine-tuned model as a candidate before trying trimming the next layer. So in this case, only one layer is trimmed in each candidate model and in all we generate $k$ candidates that satisfy the sub-target in one iteration.

Finally, we select the model with the highest accuracy from the candidates. Therefore, in one iteration, we essentially trim only one layer with minimum accuracy loss to satisfy the sub-target $T_{subi}$.

Above is the trimming approach we use in our CompactNet. We need to emphasize that it is not the only option to do it. We can also consider other methods not only for choosing which filters remaining in each layer as mentioned but also for the whole trimming approach. For example, we can use other pruning approaches like reinforcement learning to trim the model more fine-grained instead of reducing the entire filters. In this point of view, our searching loop (see Figure 1 and Figure 3) is more like a general algorithm framework that makes various trimming approaches become platform-aware and automatic.

\section{Experiments}

We choose the state-of-the-art slim CNN model MobileNetV2~\cite{18} as our target model for optimizing, and we agree with the viewpoint in~\cite{30} that traditional CNN models are designed redundantly so it is easy to achieve great optimization on them. So the more significant challenge should be optimizing the models which have already achieved a great tradeoff between speed and accuracy.

We experiment on a HUAWEI Mate10 smartphone and apply our CompactNet on two processors, an ARM mobile CPU and a domain-specific accelerator called NPU with Cambricon-1A ISA~\cite{26}. Both processors are typical backend hardware platform for CNNs applications in mobile devices.

We use the Cifar-10 dataset~\cite{31} with 50K images as training data and another 10K as validation data to pre-train an original MobileNetV2 model. Then driven by the simulated latency data of the two platforms in the smartphone, we implement the searching loop to optimize the pre-trained model for the two backend platforms. We also validate our platform simulator which is the foundation of our work to see whether it can simulate the latency precisely. The details of the experiments are discussed in this section. Note that the code of our work has been released and all the experiment results can be easily reproduced.

\subsection{Configurations}

As described in Section 3.2, driven by the simulated latency data, the searching loop (see Figure 3) does not need to be deployed or interact with the actual backend platform. So we implement the algorithm on a desktop with an Nvidia Geforce GTX10801Ti GPU and an Intel Xeon E5-2620v3 CPU since we have CNN training workloads here. The searching loop generating the optimal model has several hyperparameters to set. These hyperparameters can be divided into two sets.

\subsubsection{The Searching Algorithm}

As described in Section 3.2, the whole searching  process is broken into N iterations and $T_{subi}$ in Eq.1 is determined by not only $T_{init}$ but $N$ and $d$ as well. $N$ should not be too small since that makes $T_{subi}$ so big that might cause losing imperative feature in layers during the trimming step. The decay is just like that of the learning rate in DL training. In our experiments, these hyperparameters are set as in Table 3. Due to the much better performance of the NPU accelerator, its optimizing space should be smaller. As a result, we lower the target speedup (both $T_{init}$ and $T_{final}$) and increase the number of iterations $N$ on the NPU platform to trim more carefully.

\begin{table}
\centering
\begin{tabular}{lrrrr}  
\toprule
Platform  &  $T_{init}$  &  $ N$  &  $d$  &  $T_{final}$    \\
\midrule
Mobile CPU       &  2.0x  &  24  &  0.96  &  1.5x      \\
                         &  2.5x  &  30  &  0.98  &  1.8x      \\
NPU Accelerator       &  1.5x  &  36  &  0.96  &  1.3x      \\
                                 &  2.0x  &  48  &  0.98  &  1.5x      \\
\bottomrule
\end{tabular}
\caption{Algorithm parameters (see Eq.1 and Eq.2) set for CPU and NPU platforms.}
\end{table}

\subsubsection{CNN Training}

There are two training processes during the searching. One is a short-term fine-tuning after trimming each layer and the other is a long-term retraining after getting the optimal model which satisfies the target speedup. Still, we use the same Cifar-10 dataset for the pre-trained model and the standard RMSPropOptimizer built in TensorFlow. The training parameters for the two processes are shown in Table 4. We use a larger learning rate with faster decay in the short-term fine-tuning to quickly converge the loss and restore the accuracy after removing filters in a certain layer of the model.

\begin{table}
\centering
\begin{tabular}{lrrrr}  
\toprule
Process  &  learning\_rate  &  decay  &  epochs  &  batch\_size    \\
\midrule
Fine-tune       &  0.001  &  0.9  &  30  &  96      \\
Retraining       &  0.0001  &  0.95  &  500  &  96      \\
\bottomrule
\end{tabular}
\caption{Training parameters set for the short-term fine-tune and the long-term retraining.}
\end{table}

\subsection{Optimizing Results}

With the configurations set as the above, CompactNet can generate a platform-specific optimal model and achieve better speedup results than other optimizing approaches on both mobile CPU and the NPU accelerator. On the other hand, CompactNet maintains accuracy with such speedup and can even slightly improve it if the speedup target is less aggressive.

\subsubsection{Mobile CPU}

We set two different speedup targets and the results are shown in Table 5. Our CompactNet achieves higher accuracy with a 1.5x speedup compared with the original MobileNetV2 and maintains the accuracy with up to a 1.8x speedup. We also compare with some state-of-the-art counterparts, including NetAdapt~\cite{16}, MorphNet~\cite{29} and ADC~\cite{15} on the same Cifar-10 dataset and our CompactNet outperforms them in terms of both speedup and accuracy than those works. Compared with the original MobileNetV2 model, the number of filters in each layer of the optimal models are shown in Figure 4.

\begin{table}
\centering
\begin{tabular}{lrc}
\toprule
Optimizing Approach  &  $T_{final}$  &  Top-1 \\ 
 & &  Accuracy (\%)    \\
\midrule
Original MobileNetV2 (100\%)       &  1.0x  &  71.98      \\
NetAdapt~\cite{16}                                      &  1.3x  &  70.63      \\
MorphNet~\cite{29}                                      &  1.2x  &  69.87      \\
ADC~\cite{15}                                               &  1.2x  &  69.51      \\
\textbf{CompactNet on mobile CPU}           &  \textbf{1.5x}  &  \textbf{72.12}      \\
                                                      &  \textbf{1.8x}  &  \textbf{71.59}      \\
\bottomrule
\end{tabular}
\caption{Final speedup $T_{final}$ and accuracy results of CompactNet deployed on \textbf{mobile CPU} compared with the original MobileNetV2 and other works.}
\end{table}

\begin{figure}
\centering
\centerline{\includegraphics[width=\columnwidth]{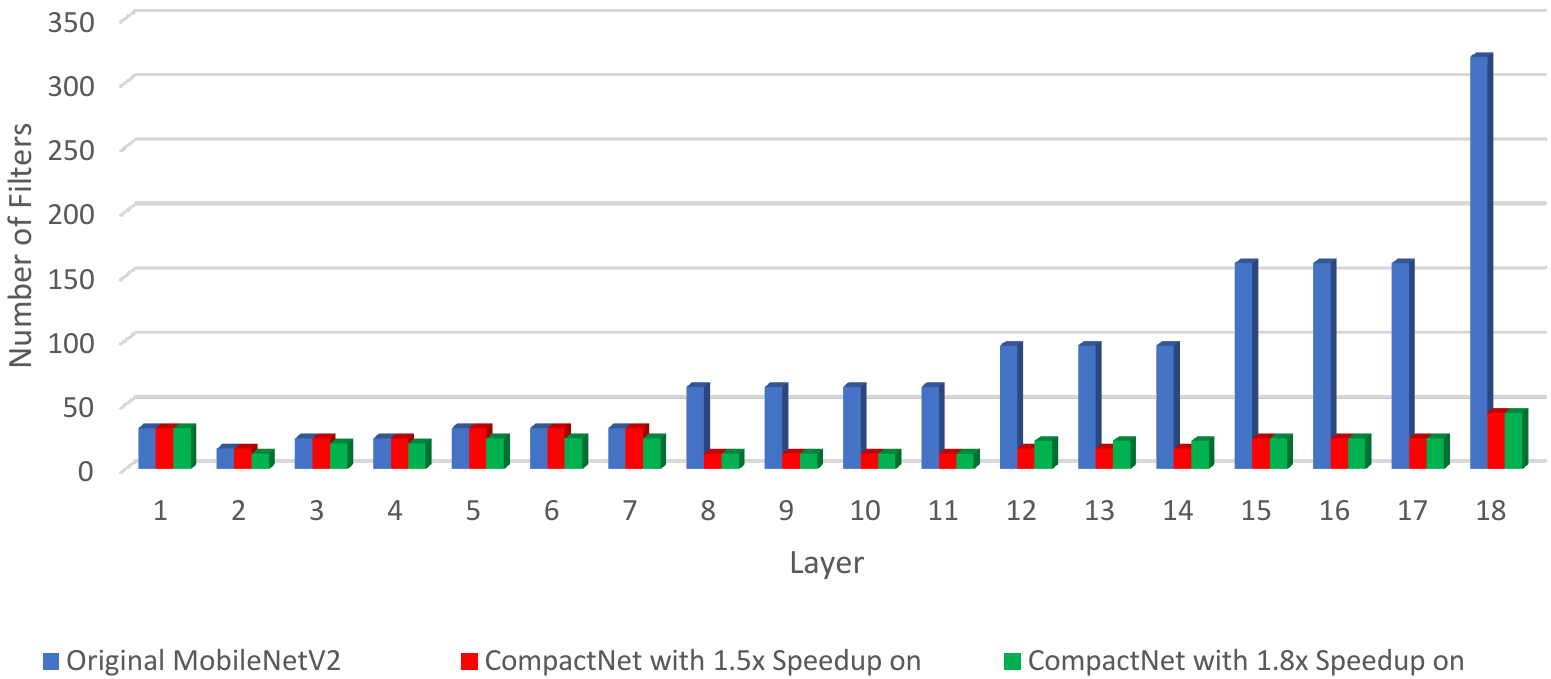}}
\caption{Number of filters per layer in the original MobileNetV2 and optimal models for \textbf{mobile CPU} generated by CompactNet.}
\end{figure}	

\subsubsection{NPU Accelerator}

Since other works are not specifically designed for the NPU platform, they cannot achieve the same speedup as that on the mobile CPU platform. In contrast, our CompactNet can still generate optimal models for the NPU. The results are shown in Table 6 and the number of filters per layer in optimal models compared with the original MobileNetV2 are shown in Figure 5.

\begin{table}
\centering
\begin{tabular}{lrc}
\toprule
Optimizing Approach  &  $T_{final}$  &  Top-1 \\ 
 & &  Accuracy (\%)    \\
\midrule
Original MobileNetV2 (100\%)       &  1.0x  &  71.98      \\
NetAdapt~\cite{16}                                      &  1.2x  &  70.63      \\
MorphNet~\cite{29}                                      &  1.1x  &  69.87      \\
ADC~\cite{15}                                               &  1.1x  &  69.51      \\
\textbf{CompactNet on NPU}           &  \textbf{1.3x}  &  \textbf{72.56}      \\
                                                      &  \textbf{1.5x}  &  \textbf{71.68}      \\
\bottomrule
\end{tabular}
\caption{Final speedup $T_{final}$ and accuracy results of CompactNet deployed on the \textbf{NPU accelerator} compared with the original MobileNetV2 and other works.}
\end{table}

\begin{figure}
\centering
\centerline{\includegraphics[width=\columnwidth]{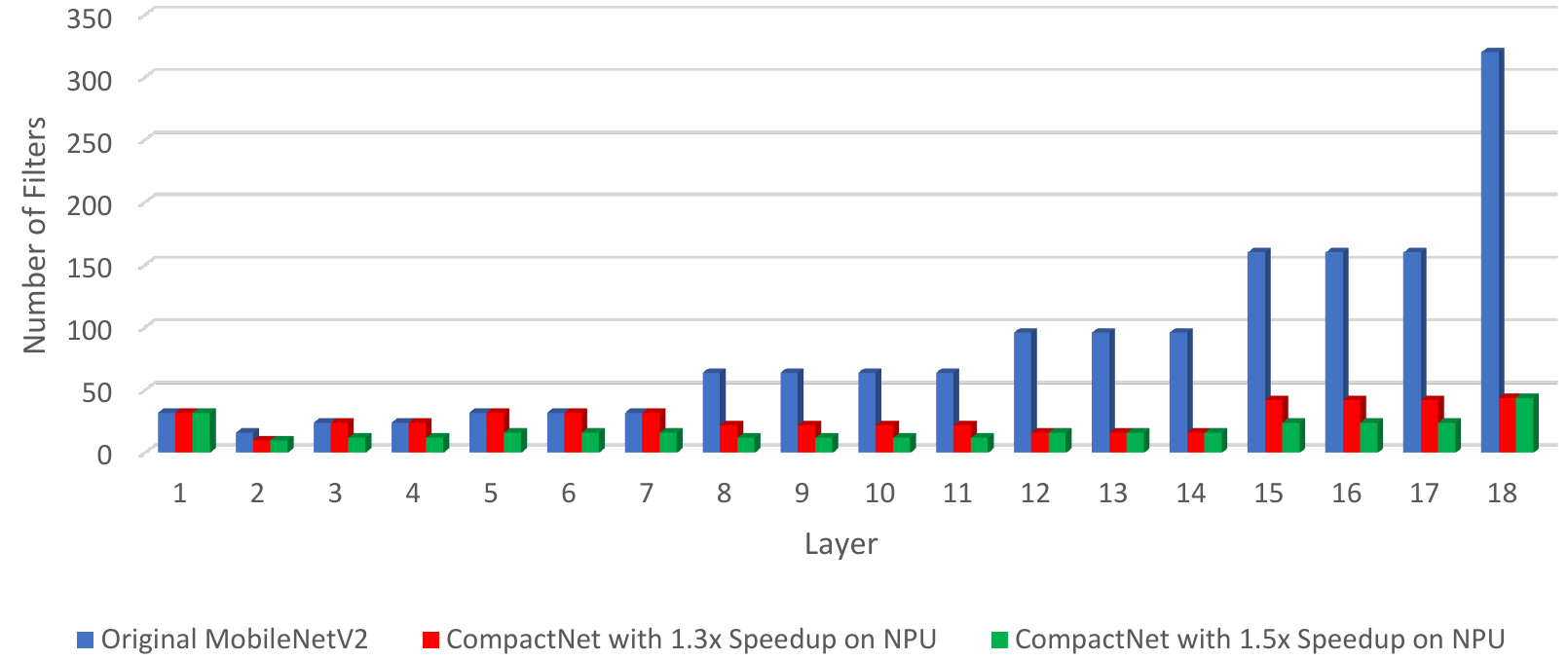}}
\caption{Number of filters per layer in original MobileNetV2 and optimal models for the \textbf{NPU accelerator} generated by CompactNet.}
\end{figure}

\

\noindent These results can also give us some interesting insight about the CNN model itself. For example, the original MobileNetV2 has an incremental-filters architecture among the layers. However, the majority of the filters trimmed by our CompactNet are in the last several layers, which might mean that many filters in those layers are redundant for the classification task on Cifar-10.

\subsubsection{Platform-Specific Models}

As in Table 5 and Table 6, we achieve a 1.5x speedup on both the mobile CPU and the NPU accelerator platforms. Regarding the number of filters per layer (see Figure 6), the optimal models for the two platforms with the same speedup are different in some layers. Furthermore, if we exchange the models between them, neither the mobile CPU nor the NPU accelerator can satisfy the target speedup (see Table 7). It suggests that the optimal model generated by our CompactNet is platform-specific. The intrinsic reason for this should be deep in the hardware architecture characteristics and our work can be considered to provide a black-box to interact with them.

\begin{figure}
\centering
\centerline{\includegraphics[width=\columnwidth]{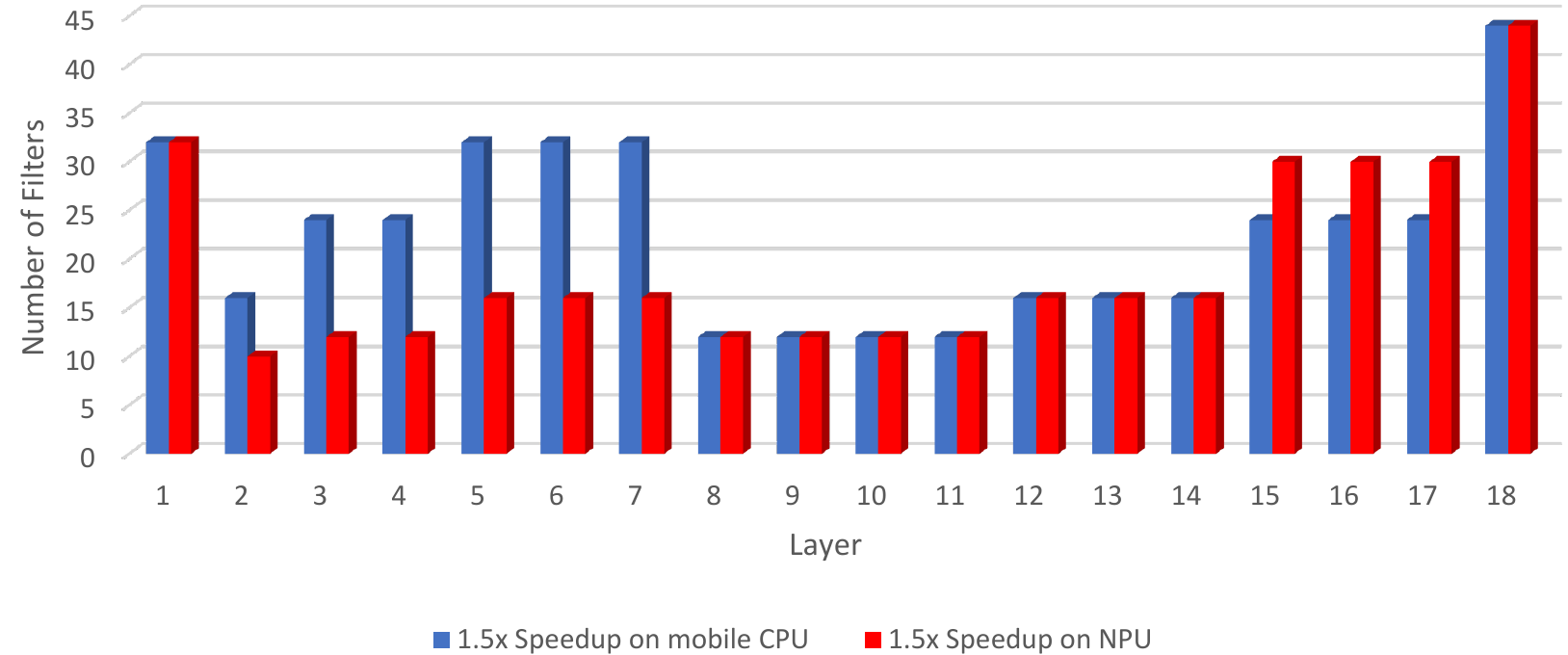}}
\caption{Comparison of optimal models of different platforms with the same speedup.}
\end{figure}

\begin{table}
\centering
\begin{tabular}{lrrc}
\toprule
Platform  &  Optimal Model  &  $T_{final}$  &  Top-1 \\ 
 & &    &  Accuracy (\%)    \\
\midrule
Mobile CPU  &  Mobile CPU   &  1.5x  &  72.12      \\
                    &  NPU             &  1.3x   &  71.68     \\
\midrule
NPU              &  NPU            &  1.5x  &  71.68      \\
                    & Mobile CPU   &  1.2x  &  72.12      \\
\bottomrule
\end{tabular}
\caption{Result of exchanging the optimal models of the two platforms. A significant drop of final speedup $T_{final}$ can be found, which means the optimal models are platform-specific}
\end{table}

\subsection{Validation of  Platform Simulator}

We use the Cambricon NeuWare SDK with Google TensorFlow~\cite{28} to implement kernel computations in each layer of the original MobileNetV2 model on both mobile CPU and the NPU accelerator to simulate the latency data. Then we compare the latency of the original model with that simulated by accumulating the latency of each layer (method shown in Figure 2). The results are shown in Figure 7. For both backend platforms, no matter how many layers are executed, the difference between the real latency and the simulated one is sufficiently small.

\begin{figure}
\centering
\centerline{\includegraphics[width=\columnwidth]{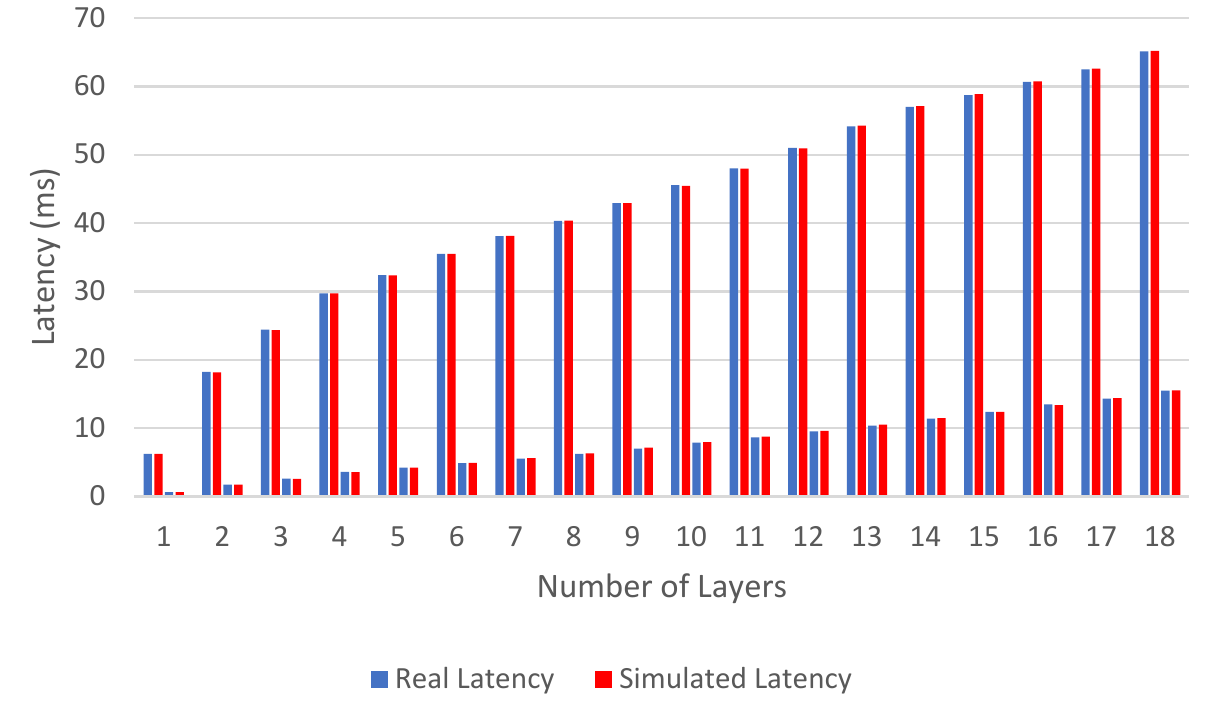}}
\caption{Kernel computation latency of executing certain layers of MobileNetV2 on real mobile CPU (longer), NPU (shorter) hardware and that simulated by the simulator. They are closed enough to validate our simulator.}
\end{figure}

\section{Conclusion}

In summary, we propose an optimizing approach, named CompactNet, to optimize a pre-trained CNN model specifically for different backend platforms given a specific target of speedup. Based on a platform simulator, CompactNet can automatically trim a pre-trained model and generate a platform-specific optimal model which satisfies the target speedup. Driven by the simulated latency data of the backend platform, CompactNet progressively trims the pre-trained model by removing the redundant filters across the layers while maintaining the accuracy by fine-tuning and retraining processes. We deploy CompactNet on a smartphone with backend platforms of a mobile CPU and a domain-specific accelerator. Compared with the state-of-the-art slim CNN model – MobileNetV2, we achieve up to a 1.8x kernel computation speedup on different platforms with equal or even higher accuracy for image classification task on the Cifar-10 dataset.

\nocite{1}
\nocite{2}
\nocite{3}
\nocite{4}
\nocite{5}
\nocite{6}
\nocite{7}
\nocite{8}
\nocite{9}
\nocite{10}
\nocite{11}
\nocite{12}
\nocite{13}
\nocite{14}
\nocite{15}
\nocite{16}
\nocite{17}
\nocite{18}
\nocite{19}
\nocite{20}
\nocite{21}
\nocite{22}
\nocite{23}
\nocite{24}
\nocite{25}
\nocite{26}
\nocite{28}
\nocite{29}
\nocite{30}
\nocite{31}

\bibliographystyle{named}
\bibliography{ijcai19}

\begin{thebibliography}{}

\bibitem[\protect\citeauthoryear{Abadi \bgroup \em et al.\egroup }{2016}]{28}
Mart{\'{\i}}n Abadi, Paul Barham, Jianmin Chen, Zhifeng Chen, Andy Davis,
  et~al.
\newblock Tensorflow: {A} system for large-scale machine learning.
\newblock In {\em 12th {USENIX} Symposium on Operating Systems Design and
  Implementation, {OSDI} 2016, Savannah, GA, USA, November 2-4, 2016.}, pages
  265--283, 2016.

\bibitem[\protect\citeauthoryear{Anwar \bgroup \em et al.\egroup }{2017}]{8}
Sajid Anwar, Kyuyeon Hwang, and Wonyong Sung.
\newblock Structured pruning of deep convolutional neural networks.
\newblock {\em {JETC}}, 13(3):32:1--32:18, 2017.

\bibitem[\protect\citeauthoryear{Cheng \bgroup \em et al.\egroup }{2018}]{1}
Jian Cheng, Peisong Wang, Gang Li, Qinghao Hu, and Hanqing Lu.
\newblock Recent advances in efficient computation of deep convolutional neural
  networks.
\newblock {\em Frontiers of {IT} {\&} {EE}}, 19(1):64--77, 2018.

\bibitem[\protect\citeauthoryear{Courbariaux and Bengio}{2016}]{23}
Matthieu Courbariaux and Yoshua Bengio.
\newblock Binarynet: Training deep neural networks with weights and activations
  constrained to +1 or -1.
\newblock {\em CoRR}, abs/1602.02830, 2016.

\bibitem[\protect\citeauthoryear{Gong \bgroup \em et al.\egroup }{2014}]{24}
Yunchao Gong, Liu Liu, Ming Yang, and Lubomir~D. Bourdev.
\newblock Compressing deep convolutional networks using vector quantization.
\newblock {\em CoRR}, abs/1412.6115, 2014.

\bibitem[\protect\citeauthoryear{Gordon \bgroup \em et al.\egroup }{2018}]{29}
Ariel Gordon, Elad Eban, Ofir Nachum, Bo~Chen, Hao Wu, et~al.
\newblock Morphnet: Fast {\&} simple resource-constrained structure learning of
  deep networks.
\newblock In {\em 2018 {IEEE} Conference on Computer Vision and Pattern
  Recognition, {CVPR} 2018, Salt Lake City, UT, USA, June 18-22, 2018}, pages
  1586--1595, 2018.

\bibitem[\protect\citeauthoryear{Guo \bgroup \em et al.\egroup }{2016}]{5}
Yiwen Guo, Anbang Yao, and Yurong Chen.
\newblock Dynamic network surgery for efficient dnns.
\newblock In {\em Advances in Neural Information Processing Systems 29: Annual
  Conference on Neural Information Processing Systems 2016, December 5-10,
  2016, Barcelona, Spain}, pages 1379--1387, 2016.

\bibitem[\protect\citeauthoryear{Han \bgroup \em et al.\egroup }{2015a}]{6}
Song Han, Huizi Mao, and William~J. Dally.
\newblock Deep compression: Compressing deep neural network with pruning,
  trained quantization and huffman coding.
\newblock {\em CoRR}, abs/1510.00149, 2015.

\bibitem[\protect\citeauthoryear{Han \bgroup \em et al.\egroup }{2015b}]{2}
Song Han, Jeff Pool, John Tran, and William~J. Dally.
\newblock Learning both weights and connections for efficient neural network.
\newblock In {\em Advances in Neural Information Processing Systems 28: Annual
  Conference on Neural Information Processing Systems 2015, December 7-12,
  2015, Montreal, Quebec, Canada}, pages 1135--1143, 2015.

\bibitem[\protect\citeauthoryear{He and Han}{2018}]{15}
Yihui He and Song Han.
\newblock {ADC:} automated deep compression and acceleration with reinforcement
  learning.
\newblock {\em CoRR}, abs/1802.03494, 2018.

\bibitem[\protect\citeauthoryear{He \bgroup \em et al.\egroup }{2017}]{12}
Yihui He, Xiangyu Zhang, and Jian Sun.
\newblock Channel pruning for accelerating very deep neural networks.
\newblock In {\em {IEEE} International Conference on Computer Vision, {ICCV}
  2017, Venice, Italy, October 22-29, 2017}, pages 1398--1406, 2017.

\bibitem[\protect\citeauthoryear{Howard \bgroup \em et al.\egroup }{2017}]{17}
Andrew~G. Howard, Menglong Zhu, Bo~Chen, Dmitry Kalenichenko, Weijun Wang,
  et~al.
\newblock Mobilenets: Efficient convolutional neural networks for mobile vision
  applications.
\newblock {\em CoRR}, abs/1704.04861, 2017.

\bibitem[\protect\citeauthoryear{Iandola \bgroup \em et al.\egroup }{2016}]{19}
Forrest~N. Iandola, Matthew~W. Moskewicz, Khalid Ashraf, Song Han, William~J.
  Dally, and Kurt Keutzer.
\newblock Squeezenet: Alexnet-level accuracy with 50x fewer parameters and
  {\textless}1mb model size.
\newblock {\em CoRR}, abs/1602.07360, 2016.

\bibitem[\protect\citeauthoryear{Jacob \bgroup \em et al.\egroup }{2018}]{30}
Benoit Jacob, Skirmantas Kligys, Bo~Chen, Menglong Zhu, Matthew Tang, et~al.
\newblock Quantization and training of neural networks for efficient
  integer-arithmetic-only inference.
\newblock In {\em 2018 {IEEE} Conference on Computer Vision and Pattern
  Recognition, {CVPR} 2018, Salt Lake City, UT, USA, June 18-22, 2018}, pages
  2704--2713, 2018.

\bibitem[\protect\citeauthoryear{Jaderberg \bgroup \em et al.\egroup
  }{2014}]{21}
Max Jaderberg, Andrea Vedaldi, and Andrew Zisserman.
\newblock Speeding up convolutional neural networks with low rank expansions.
\newblock In {\em British Machine Vision Conference, {BMVC} 2014, Nottingham,
  UK, September 1-5, 2014}, 2014.

\bibitem[\protect\citeauthoryear{Kim \bgroup \em et al.\egroup }{2015}]{22}
Yong{-}Deok Kim, Eunhyeok Park, Sungjoo Yoo, Taelim Choi, Lu~Yang, and Dongjun
  Shin.
\newblock Compression of deep convolutional neural networks for fast and low
  power mobile applications.
\newblock {\em CoRR}, abs/1511.06530, 2015.

\bibitem[\protect\citeauthoryear{Krizhevsky and Hinton}{2009}]{31}
A~Krizhevsky and G~Hinton.
\newblock Learning multiple layers of features from tiny images.
\newblock {\em Computer Science Department, University of Toronto, Tech. Rep},
  1, 01 2009.

\bibitem[\protect\citeauthoryear{Lebedev and Lempitsky}{2016}]{9}
Vadim Lebedev and Victor~S. Lempitsky.
\newblock Fast convnets using group-wise brain damage.
\newblock In {\em 2016 {IEEE} Conference on Computer Vision and Pattern
  Recognition, {CVPR} 2016, Las Vegas, NV, USA, June 27-30, 2016}, pages
  2554--2564, 2016.

\bibitem[\protect\citeauthoryear{Liu \bgroup \em et al.\egroup }{2016}]{26}
Shaoli Liu, Zidong Du, Jinhua Tao, Dong Han, Tao Luo, Yuan Xie, Yunji Chen, and
  Tianshi Chen.
\newblock Cambricon: An instruction set architecture for neural networks.
\newblock In {\em 43rd {ACM/IEEE} Annual International Symposium on Computer
  Architecture, {ISCA} 2016, Seoul, South Korea, June 18-22, 2016}, pages
  393--405, 2016.

\bibitem[\protect\citeauthoryear{Luo \bgroup \em et al.\egroup }{2017}]{13}
Jian{-}Hao Luo, Jianxin Wu, and Weiyao Lin.
\newblock Thinet: {A} filter level pruning method for deep neural network
  compression.
\newblock In {\em {IEEE} International Conference on Computer Vision, {ICCV}
  2017, Venice, Italy, October 22-29, 2017}, pages 5068--5076, 2017.

\bibitem[\protect\citeauthoryear{Mao \bgroup \em et al.\egroup }{2017}]{7}
Huizi Mao, Song Han, Jeff Pool, Wenshuo Li, Xingyu Liu, et~al.
\newblock Exploring the regularity of sparse structure in convolutional neural
  networks.
\newblock {\em CoRR}, abs/1705.08922, 2017.

\bibitem[\protect\citeauthoryear{Molchanov \bgroup \em et al.\egroup
  }{2016}]{4}
Pavlo Molchanov, Stephen Tyree, Tero Karras, Timo Aila, and Jan Kautz.
\newblock Pruning convolutional neural networks for resource efficient transfer
  learning.
\newblock {\em CoRR}, abs/1611.06440, 2016.

\bibitem[\protect\citeauthoryear{Sandler \bgroup \em et al.\egroup }{2018}]{18}
Mark Sandler, Andrew~G. Howard, Menglong Zhu, Andrey Zhmoginov, and
  Liang{-}Chieh Chen.
\newblock Mobilenetv2: Inverted residuals and linear bottlenecks.
\newblock In {\em 2018 {IEEE} Conference on Computer Vision and Pattern
  Recognition, {CVPR} 2018, Salt Lake City, UT, USA, June 18-22, 2018}, pages
  4510--4520, 2018.

\bibitem[\protect\citeauthoryear{Wen \bgroup \em et al.\egroup }{2016}]{10}
Wei Wen, Chunpeng Wu, Yandan Wang, Yiran Chen, and Hai Li.
\newblock Learning structured sparsity in deep neural networks.
\newblock In {\em Advances in Neural Information Processing Systems 29: Annual
  Conference on Neural Information Processing Systems 2016, December 5-10,
  2016, Barcelona, Spain}, pages 2074--2082, 2016.

\bibitem[\protect\citeauthoryear{Yang \bgroup \em et al.\egroup }{2017}]{25}
Tien{-}Ju Yang, Yu{-}Hsin Chen, and Vivienne Sze.
\newblock Designing energy-efficient convolutional neural networks using
  energy-aware pruning.
\newblock In {\em 2017 {IEEE} Conference on Computer Vision and Pattern
  Recognition, {CVPR} 2017, Honolulu, HI, USA, July 21-26, 2017}, pages
  6071--6079, 2017.

\bibitem[\protect\citeauthoryear{Yang \bgroup \em et al.\egroup }{2018}]{16}
Tien{-}Ju Yang, Andrew~G. Howard, Bo~Chen, Xiao Zhang, Alec Go, et~al.
\newblock Netadapt: Platform-aware neural network adaptation for mobile
  applications.
\newblock In {\em Computer Vision - {ECCV} 2018 - 15th European Conference,
  Munich, Germany, September 8-14, 2018, Proceedings, Part {X}}, pages
  289--304, 2018.

\bibitem[\protect\citeauthoryear{Yu \bgroup \em et al.\egroup }{2017}]{11}
Jiecao Yu, Andrew Lukefahr, David~J. Palframan, Ganesh~S. Dasika, Reetuparna
  Das, et~al.
\newblock Scalpel: Customizing {DNN} pruning to the underlying hardware
  parallelism.
\newblock In {\em Proceedings of the 44th Annual International Symposium on
  Computer Architecture, {ISCA} 2017, Toronto, ON, Canada, June 24-28, 2017},
  pages 548--560, 2017.

\bibitem[\protect\citeauthoryear{Zhang \bgroup \em et al.\egroup }{2018}]{20}
Xiangyu Zhang, Xinyu Zhou, Mengxiao Lin, and Jian Sun.
\newblock Shufflenet: An extremely efficient convolutional neural network for
  mobile devices.
\newblock In {\em 2018 {IEEE} Conference on Computer Vision and Pattern
  Recognition, {CVPR} 2018, Salt Lake City, UT, USA, June 18-22, 2018}, pages
  6848--6856, 2018.

\end{thebibliography}

\end{document}